\documentclass[10pt,letterpaper,twocolumn,teaser]{planstyle}
\usepackage[T1]{fontenc}
\usepackage{microtype}           
\usepackage{nicefrac}            
\usepackage{xspace}              
\usepackage{amsmath, amssymb, amsfonts, amsthm, mathtools, mathrsfs}
\usepackage{dsfont}              
\usepackage{fdsymbol}            
\DeclareSymbolFont{cmmathsymbols}{OMS}{cmsy}{m}{n}
\DeclareMathSymbol{\cmnabla}{\mathord}{cmmathsymbols}{"72}
\let\nabla\cmnabla

\usepackage{booktabs}            
\usepackage{nicematrix}         
\usepackage{multirow}
\usepackage{makecell}
\usepackage{bigstrut}
\usepackage{enumitem}            
\setitemize{label=\textbullet, leftmargin=*, nolistsep}
\usepackage{graphicx}
\usepackage{adjustbox}           
\usepackage{float}               
\usepackage{wrapfig}             
\usepackage{subcaption}          
\expandafter\def\csname ver@subfig.sty\endcsname{}  
\usepackage[dvipsnames]{xcolor}    
\usepackage{color}
\usepackage{bxcoloremoji}        
\usepackage{fontawesome5}        
\usepackage{pifont}              
\usepackage{hyperref}            
\hypersetup{colorlinks=true, citecolor=LightBlue}
\usepackage[all]{hypcap}         
\usepackage{cleveref}            
\usepackage{url}                 
\usepackage{algorithm}
\usepackage{algorithmicx}
\usepackage{algpseudocode}
\usepackage[normalem]{ulem}      
\usepackage{lipsum}              
\usepackage{rotating}            
\usepackage{stackengine}         
\usepackage{caption}             
\usepackage{listings}            
\usepackage{soul}                
\usepackage{gradient-text}       
\usepackage{pgfplots}            
\pgfplotsset{compat=newest}
\usepackage{pgfplotstable}       
\usepackage{tikz}                
\usepackage{svg}                 

\usepackage[comma,numbers,sort,compress]{natbib}
\definecolor{demphcolor}{RGB}{125,125,125}    

\hypersetup{
    colorlinks = true,
    citecolor = {YaleBlue},
}
\tcbuselibrary{breakable, skins}
\newtcolorbox{planbox}[1]{
  enhanced,
  breakable,
  colback=white,
  colframe=IllinoisOrange!80,
  coltitle=IllinoisBlue,
  fonttitle=\bfseries\sffamily,
  title=#1,
  titlerule=0.8pt,
  boxrule=1pt,
  left=3mm, right=3mm, top=2mm, bottom=2mm,
  boxsep=1mm,
  before upper=\smallskip,
}

\newcommand{\ie}{\textit{i.e.},\xspace}      
\newcommand{\eg}{\textit{e.g.},\xspace}      

\definecolor{VtamV}{HTML}{2A5487}  
\definecolor{VtamM}{HTML}{13294B}  
\newcommand{\vtamc}{\textcolor{VtamV}{V}\textcolor{IllinoisOrange}{TA}\textcolor{VtamM}{M}\xspace}  
\newcommand{\vtam}{VTAM\xspace}  

\crefname{equation}{Eq.}{Eqs.}
\crefformat{section}{\S#2#1#3}
\crefformat{subsection}{\S#2#1#3}
\crefformat{subsubsection}{\S#2#1#3}
\crefrangeformat{section}{\S\S#3#1#4 to~#5#2#6}
\crefmultiformat{section}{\S\S#2#1#3}{ and~#2#1#3}{, #2#1#3}{ and~#2#1#3}

\usetikzlibrary{calc}

\title{\texorpdfstring{\vtamc}{VTAM}: Efficient Visuo-Tactile World-Action Models for Contact-Rich Manipulation without Tactile Pretraining}

\author{
\textbf{
Haoran Yuan\textsuperscript{1,*,\ddagger},
Weigang Yi\textsuperscript{1,*},
Zhenyu Zhang\textsuperscript{2,*},
Wendi Chen\textsuperscript{3,*}
}\\
Yuchen Mo\textsuperscript{1}, Jiashi Yin\textsuperscript{1}, Xinzhuo Li\textsuperscript{1}, Yue Luo\textsuperscript{1}, Jialin Li\textsuperscript{1}, Xiangyu Zeng\textsuperscript{1} \\
Chuan Wen\textsuperscript{3},
Cewu Lu\textsuperscript{3},
Katherine Driggs-Campbell\textsuperscript{1},
Ismini Lourentzou\textsuperscript{1,\dagger}
}

\affil{
\textsuperscript{1}University of Illinois Urbana-Champaign\quad
\textsuperscript{2}Stanford University\quad
\textsuperscript{3}Shanghai Jiao Tong University
}

\correspondingauthor{
\textsuperscript{\ddagger}Project Lead.
\textsuperscript{*}Equal contribution.
 \textsuperscript{\dagger}Corresponding author.
}

\renewcommand{\insertteaserfigure}{
  \begin{center}
    \vspace{-0.5cm}
    \centering
    \includegraphics[width=\textwidth, trim=18 7 8 47, clip]{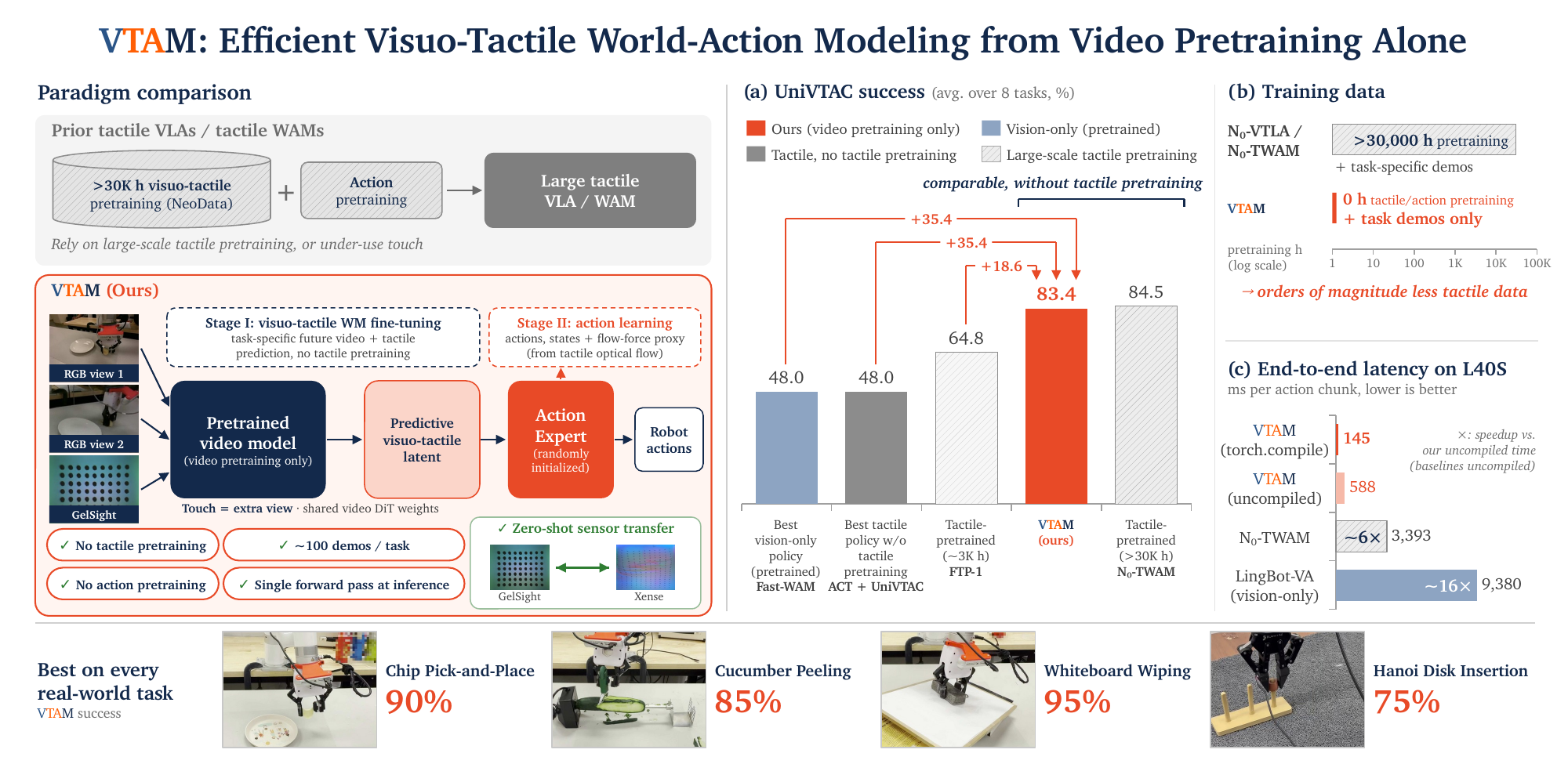}
    \captionof{figure}{\textbf{\vtamc: efficient visuo-tactile world-action modeling without tactile pretraining.} With only task demonstrations and no tactile or action pretraining, \vtam outperforms the best vision-only and non-tactile-pretrained tactile policies, is comparable to N$_0$-TWAM (${>}30$K\,h of tactile pretraining), and skips full video denoising for fast inference.}
    \label{fig:teaser}
  \end{center}
}

\begin{document}
\setabstractlogo[9mm]{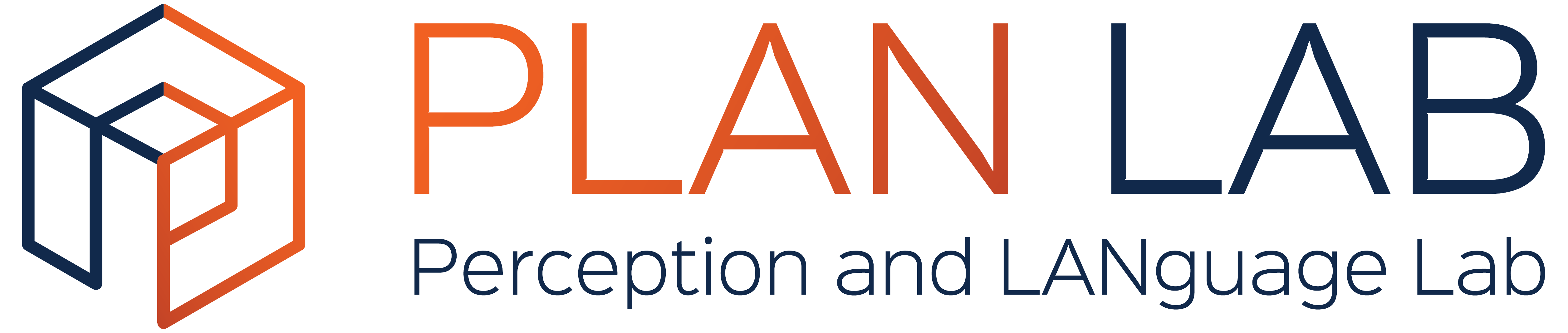}

\begin{abstract}
Tactile vision-language-action models (VLAs) and tactile world-action models (WAMs) are a promising new paradigm for contact-rich manipulation, but existing models either underutilize touch, rely on large-scale tactile pretraining, or incur high inference latency due to iterative video generation. We present \vtam, an efficient visuo-tactile world-action model for contact-rich manipulation that requires no tactile or action pretraining. RGB and tactile observations share a frozen video VAE and the pretrained video transformer weights. \vtam first adapts the video backbone to jointly predict future visual and tactile latents, then fine-tunes it end-to-end with a randomly initialized action expert that predicts actions, proprioceptive states, and a light flow-force proxy computed from tactile optical flow.
Trained only on task demonstrations, \vtam outperforms all baselines on four real-robot tasks across two platforms with different tactile sensors, outperforms all baselines without tactile pretraining in average success on eight UniVTAC simulation tasks, and is comparable to tactile VLAs and WAMs pretrained on over 30K hours of largely visuo-tactile robot data (83.4\% vs.\ 84.5\% for N$_0$-TWAM).
Without full video denoising at inference, \vtam is approximately 6$\times$ and 16$\times$ faster than N$_0$-TWAM and LingBot-VA without compilation, with compiled latency comparable to Fast-WAM. Ablations show that feeding frozen-VAE tactile features directly to the action expert, bypassing tactile world modeling, fails on every tested task, that Stage~I visuo-tactile world-model fine-tuning improves UniVTAC success on most tasks, and that flow-force proxy keeps the policy from fully ignoring touch. The tactile encoder transfers zero-shot across GelSight Mini and Xense Photon sensors.
\vspace{2mm}

\url{https://plan-lab.github.io/vtam}
\vspace{-2mm}
\end{abstract}

\maketitle

\section{Introduction}
\label{sec:intro}
Successful manipulation depends not only on the robot's motion but also on the contact it establishes with the object. A gripper may follow the intended trajectory yet close above a fragile chip. Similarly, an eraser may cross a board without removing the stain. Distinguishing these motions from effective manipulation requires observing contact conditions that external cameras may not fully reveal. Optical tactile sensors complement camera views of the object and workspace by capturing local deformation at the contact interface, providing feedback about whether the intended interaction is occurring
~\cite{yuan2017gelsight,dong2017improved,yuan2015shear}.\looseness-1

Incorporating this feedback into general-purpose robot models offers a path toward more reliable contact-rich manipulation. Vision-language-action models (VLAs) learn robot behaviors from visual, linguistic, and action data~\cite{zitkovich2023rt,kim2024openvla,black2024pi_0,intelligence2025pi_}, while video-based world-action models (WAMs) use predictive representations of how interactions unfold to generate actions~\cite{liao2025genie,zhu2025unified,li2025unified}. Because optical tactile sensors produce image sequences, their observations can also be modeled within a video-based framework. This provides an opportunity to learn representations that connect changes in the visual scene with the evolution of local contact.\looseness-1

Building on tactile representations~\cite{yang2024binding,higuera2024sparsh,xu2025unit} and visuo-tactile policies~\cite{huang20243d,xue2025reactive,zhuo2026fardp}, tactile VLAs~\cite{huang2025tactile,hao2025tla,bi2025vla,wang2026never} and tactile WAMs~\cite{wu2026tactile,neoteai2026twam} have recently emerged as a new paradigm. 
However, they either use touch in a limited way, \eg as an additional input to a policy, \ie adding tactile features to representations learned without touch, or rely on large-scale tactile pretraining~\cite{yuan2026ftp,neoteai2026vtla,neoteai2026twam}, which requires costly visuo-tactile robot data. Moreover, a policy may underuse touch when richer, faster-changing visual features already explain much of the demonstration data~\cite{liu2025factr,zhang2025ta,chen2025implicitrdp}.\looseness-1

We introduce \textbf{\vtamc}, an efficient visuo-tactile world-action model for contact-rich manipulation without tactile pretraining (Fig.~\ref{fig:teaser}). \vtam adapts a pretrained video backbone, shared by RGB and tactile views, through joint prediction of future visual and tactile observations, then uses the resulting representations for action learning. Both predictive tactile modeling and action learning rely solely on task demonstrations, with no tactile or action pretraining. An auxiliary flow-force proxy derived from tactile optical flow encourages the policy to use contact information without requiring a force sensor. 

\begin{figure*}[t]
  \centering
  \includegraphics[width=0.615\textwidth]{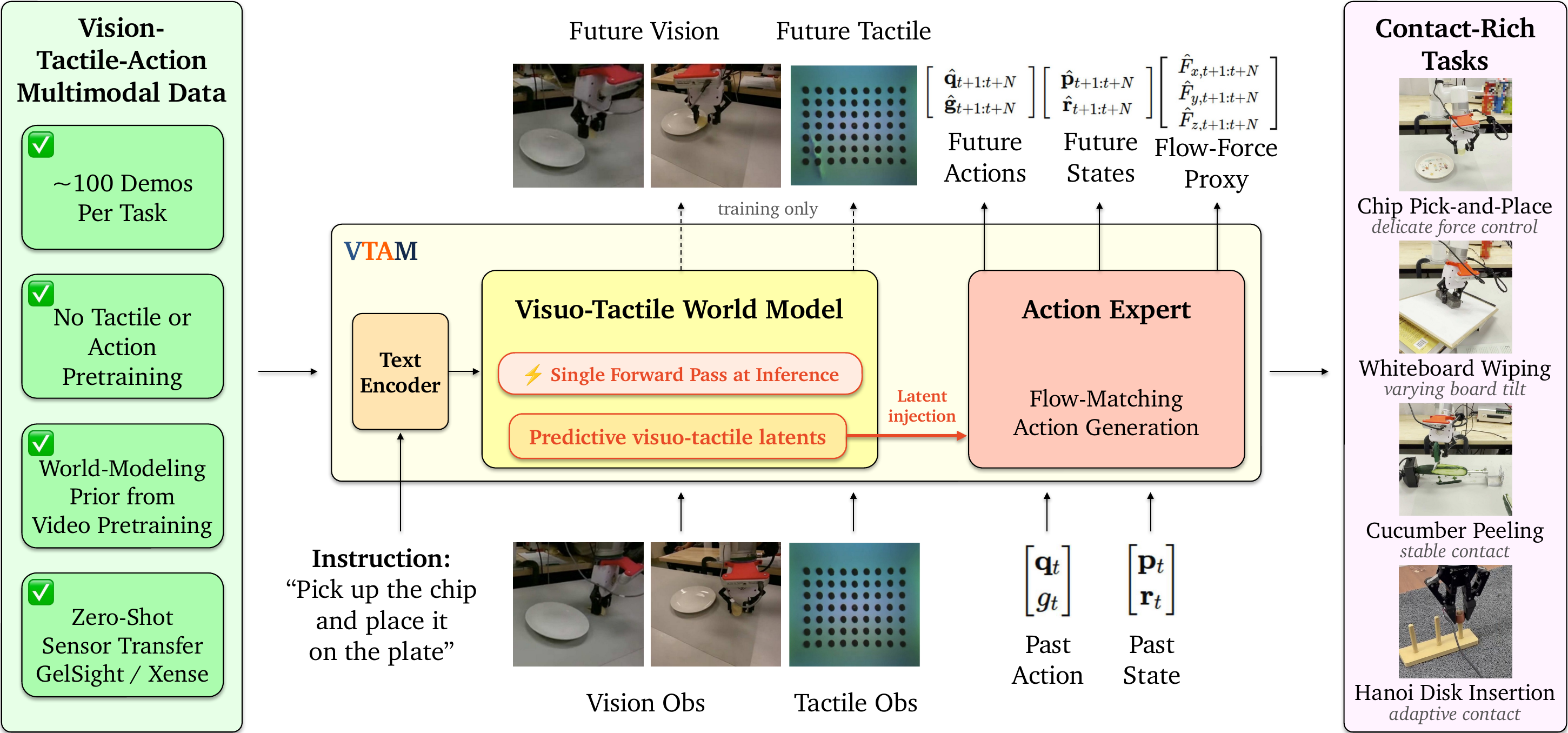}\hfill
  \includegraphics[width=0.375\textwidth]{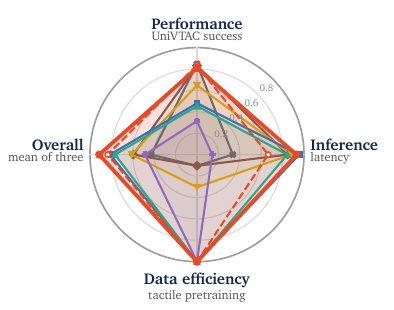}\\[2pt]
  \includegraphics[width=\textwidth]{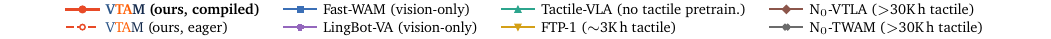}
  \caption{\textbf{\vtamc overview (left) and performance, inference efficiency, and data efficiency (right).} \vtam fine-tunes a pretrained video model into a visuo-tactile world model and runs it once per action chunk, matching tactile-pretrained models in success without tactile pretraining. Outer is better; latency and data axes are log-scaled, baselines use their fastest (eager or compiled) latency, and Overall averages the three normalized scores.}
  \label{fig:radar}
\end{figure*}
With approximately 100 demonstrations per task (Fig.~\ref{fig:radar}), \vtam achieves 75–95\% success across four real-robot tasks on two platforms with different tactile sensors, outperforming all evaluated baselines. On eight UniVTAC tasks, it reaches 83.4\% average success, outperforming all baselines without tactile pretraining (at most 48.0\%). This performance is comparable to N$_0$ models pretrained on over 30K hours of largely visuo-tactile robot data (83.1\% and 84.5\%)~\cite{neoteai2026vtla,neoteai2026twam}. Avoiding full video denoising also makes uncompiled \vtam approximately 6$\times$ faster per action chunk than N$_0$-TWAM and 16$\times$ faster than LingBot-VA, with compiled latency comparable to Fast-WAM~\cite{yuan2026fastwam}.
Ablations support the importance of both predictive tactile learning and contact-related supervision. Finally, the tactile encoder transfers zero-shot across tactile sensors, and its latents trace contact progress over time and differ across tasks. 
Our contributions are:
\begin{itemize}[itemsep=0ex, parsep=0pt, topsep=-1ex, leftmargin=0.5cm]
    \item \vtam, an efficient visuo-tactile world-action model that couples self-supervised tactile world modeling with action learning, without tactile or action pretraining and without video generation at inference. 
    \item A light flow-force proxy, computed directly from tactile images and predicted by the action expert, encouraging tactile use during action learning without an external force sensor.\looseness-1
    \item An evaluation on eight UniVTAC tasks and four real-robot tasks on two platforms (xArm6 with GelSight Mini; Flexiv Rizon 4 with Xense Photon) showing data efficiency, competitive performance and inference speed, and zero-shot cross-sensor tactile encoding.
\end{itemize}

\section{Related Work}
\label{sec:related_works}
\noindent\textbf{World-action models.}
VLAs build on pretrained vision-language models~\cite{zitkovich2023rt,kim2024openvla,black2024pi_0,intelligence2025pi_,driess2023palm,team2024octo}, and predictive policies model future observations~\cite{seo2023masked,zhen20243d,zhang2025dreamvla}. WAMs couple video prediction with action generation~\cite{liao2025genie,zhu2025unified,li2025unified}; \vtam builds on a pretrained video model~\cite{liao2025genie}. LingBot-VA~\cite{li2026lingbotva} and N$_0$-TWAM~\cite{neoteai2026twam} fully denoise future frames at inference (9.4\,s and 3.4\,s per chunk), and ForceWAM~\cite{xu2026forcewam} jointly predicts visual and force futures for reactive contact-rich manipulation. Fast-WAM~\cite{yuan2026fastwam} skips full video denoising. In contrast, \vtam uses visuo-tactile denoising only as a self-supervised training objective; at inference, one forward pass yields visuo-tactile latents for the action expert without full video denoising, at 145\,ms per chunk (compiled), on par with the vision-only Fast-WAM (127\,ms).\looseness-1

\noindent\textbf{Tactile VLAs and visuo-tactile policies.}
Visuo-tactile policies (3D-ViTac~\cite{huang20243d}, RDP~\cite{xue2025reactive}, FA-RDP~\cite{zhuo2026fardp}, UniVTAC's ACT~\cite{chen2026univtac}) and tactile or force-aware VLAs and WAMs (Tactile-VLA, TLA, VLA-Touch, AT-VLA, ForceVLA, Tactile-WAM~\cite{huang2025tactile,hao2025tla,bi2025vla,li2026atvla,yu2025forcevla,wu2026tactile,wang2026never}) add touch or force as extra inputs, yet a policy may underuse touch when richer visual features already explain the demonstrations~\cite{wang2020multimodal,wu2022characterizing,liu2025factr,zhang2025ta,chen2025implicitrdp}. FTP-1~\cite{yuan2026ftp} and N$_0$-VTLA/N$_0$-TWAM~\cite{neoteai2026vtla,neoteai2026twam} instead rely on ${\sim}$3K and over 30K hours of tactile pretraining; N$_0$-TWAM also trains a dedicated tactile expert from scratch on this data. 
In contrast, \vtam processes touch with the pretrained video DiT itself and combines tactile world modeling with flow-force proxy supervision, learning contact-sensitive control from a pretrained video model. Using only the provided UniVTAC demonstrations, it is comparable to N$_0$-TWAM (83.4\% vs.\ 84.5\%) without large-scale tactile pretraining.\looseness-1

\noindent\textbf{Tactile representations and force supervision.}
Tactile encoders use multimodal alignment~\cite{yang2024binding,dave2024multimodal}, self-supervised pretraining~\cite{higuera2024sparsh}, or data-efficient training~\cite{xu2025unit}; tactile models also address slip~\cite{parag2024learning,zapata2019learning,wang2023robust}, simulation~\cite{si2024difftactile}, and visuo-tactile prediction~\cite{higuera2026vtwm,zheng2026omnivta}. Tactile deformation reflects contact loading~\cite{yuan2015shear,sundaralingam2019robust,zhu2025forces}, but force labels are hard to obtain for a gripper: a wrist F/T sensor cancels the opposing fingertip forces, in-sensor force sensing is costly and bulky, and Sharpa trains a separate estimator~\cite{su2026tacmap}. In contrast, \vtam refines its visuo-tactile representation by a self-supervised tactile world-modeling objective rather than frozen features, uses a flow-force proxy that needs no force sensor or learned estimator, and its tactile encoder transfers zero-shot across GelSight Mini and Xense Photon.

\section{Method}
\label{sec:method}
\vtam combines a predictive visuo-tactile backbone with a conditional action expert (Fig.~\ref{fig:vtam_architecture}). At observation time $k$, the inputs comprise two RGB views $I_k^1,I_k^2$, a GelSight image $I_k^3$ (Hanoi: one RGB view, Xense Photon), and the proprioceptive state $s_k$. Stage I performs task-specific visuo-tactile world-model fine-tuning, extending the pretrained video model to jointly predict future visual and tactile latents without any tactile pretraining. Stage II initializes from Stage I and fine-tunes the backbone end-to-end with the expert to predict actions, states, and a flow-force proxy. 

\subsection{Predictive Visuo-Tactile Representation}
We initialize the representation from the GE-base checkpoint of Genie Envisioner~\cite{liao2025genie}, which uses an LTX-Video backbone~\cite{hacohen2024ltx}. We reuse its reconstruction-oriented video VAE for each sensory stream, including tactile texture and deformation, so neither a separate tactile encoder nor tactile-language pairs are needed. 

Tactile frames enter the pretrained video DiT as an additional view and share its transformer weights with the RGB views, rather than being processed by a separately trained tactile expert. Touch thus adds no modality-specific transformer parameters, and visual and tactile modeling remain parameter-coupled throughout fine-tuning. Only this video model is pretrained; touch is learned from the task demonstrations (Section~\ref{sec:experiments}).

Optical tactile images describe local surface texture and deformation, whereas the two RGB views provide object and workspace context. The shared VAE avoids first reducing touch to a semantic embedding. Stage I then learns task-specific visual--tactile relationships through future prediction before policy fitting.

For view $v\in\{1,2,3\}$, let $z_k^v=E(I_k^v)$ denote its encoded observation, with the same encoder $E$ used across streams. A sequence of $B$ attention blocks processes the corresponding latent tokens. Writing $Z_{b-1}^v$ for the tokens of stream $v$ entering block $b$, the intra-view update is
\begin{equation}
\widetilde Z_b^v=\operatorname{SelfAttention}(Z_{b-1}^v),\qquad v\in\{1,2,3\}.
\label{eq:inview}
\end{equation}
The cross-view update then exchanges information across streams:
\begin{equation}
\begin{aligned}
Z_b=\operatorname{CrossViewAttention}\bigl(
\operatorname{Concat}(&\widetilde Z_b^1,\widetilde Z_b^2,\widetilde Z_b^3)\bigr).
\end{aligned}
\label{eq:crossview}
\end{equation}
Tokens span the observation sequence, capturing within-stream evolution and cross-stream dependencies at  pretrained depth $B\!=\!28$. Thus, tactile observations participate in the predictive backbone before the action-learning stage.
\begin{figure}[!t]
    \centering
    \resizebox{0.99\columnwidth}{!}{\begin{tikzpicture}[x=1pt,y=-1pt,
  font=\fontsize{7.5}{8.5}\selectfont,
  module/.style={draw=black!70,rounded corners=2pt,align=center,
    inner sep=1.2pt,outer sep=0pt},
  arr/.style={->,>=stealth,line width=0.55pt},
  wire/.style={line width=0.55pt},
  token/.style={draw=black!65,rounded corners=1.5pt,
    minimum width=9pt,minimum height=7pt,inner sep=0pt,outer sep=0pt}]
\path[use as bounding box] (0,0) rectangle (245,234);

\node[anchor=west,font=\fontsize{8}{9}\selectfont\bfseries]
  at (12,5) {Video backbone};
\node[anchor=east,font=\fontsize{7}{8}\selectfont]
  at (231,5) {28 repeated blocks};
\node[align=center] at (43,21) {Rear RGB\\VAE latents};
\node[align=center] at (122,21) {Front RGB\\VAE latents};
\node[align=center] at (201,21) {Tactile\\VAE latents};

\foreach \x in {31,43,55}
  \node[token,fill=blue!12] at (\x,35) {};
\foreach \x in {110,122,134}
  \node[token,fill=violet!10] at (\x,35) {};
\foreach \x in {189,201,213}
  \node[token,fill=violet!25] at (\x,35) {};

\draw[rounded corners=3pt,draw=black!65,fill=cyan!3]
  (10,43) rectangle (234,122);

\node[module,fill=green!7,minimum width=208pt,minimum height=14pt]
  (vtop) at (122,51.5) {Cross-view self-attention};

\foreach \x in {43,122,201} {
  \node[module,fill=yellow!10,minimum width=64pt,minimum height=22pt]
    (view\x) at (\x,75.5) {Shared in-view\\self-attention};
  \draw[arr] (\x,38.5) -- (\x,44.5);
  \draw[arr] (\x,58.5) -- (\x,64.5);
}
\foreach \x/\y in {75/90,154/169}
  \draw[<->,>=stealth,densely dashed,line width=0.5pt,draw=black!60] (\x+0.8,75.5) -- (\y-0.8,75.5);

\node[module,fill=green!7,minimum width=208pt,minimum height=14pt]
  (vbottom) at (122,99.5) {Cross-view self-attention};

\foreach \x in {43,122,201}
  \draw[arr] (\x,86.5) -- (\x,92.5);

\draw[dashed,rounded corners=1.5pt,draw=black!55]
  (16,111) rectangle (229,121);
\foreach \x in {31,43,55}
  \node[token,fill=blue!12] at (\x,116) {};
\foreach \x in {110,122,134}
  \node[token,fill=violet!10] at (\x,116) {};
\foreach \x in {189,201,213}
  \node[token,fill=violet!25] at (\x,116) {};

\foreach \x in {43,122,201}
  \draw[arr] (\x,106.5) -- (\x,112.5);

\node[font=\fontsize{7}{8}\selectfont] at (122,129)
  {Stage I: predict future visual and tactile latents};

\node at (122,141) {Current state $c_k$ + noisy joint tokens $y_\tau$};
\foreach \x in {91,103}
  \node[token,fill=red!22] at (\x,152) {};
\foreach \x in {115,127}
  \node[token,fill=blue!22] at (\x,152) {};
\foreach \x in {139,151}
  \node[token,fill=magenta!22] at (\x,152) {};

\node[anchor=west,font=\fontsize{8}{9}\selectfont\bfseries]
  at (12,152) {Action expert};
\node[anchor=east,font=\fontsize{7}{8}\selectfont]
  at (231,152) {28 repeated blocks};

\draw[rounded corners=3pt,draw=black!65,fill=orange!3]
  (10,160) rectangle (234,200);

\node[module,fill=green!7,minimum width=208pt,minimum height=14pt]
  (atop) at (122,168.5) {Self-attention};
\node[module,fill=orange!20,minimum width=208pt,minimum height=16pt]
  (across) at (122,189.5) {Multi-view cross-attention};

\draw[arr] (122,155.5) -- (122,161.5);
\draw[arr] (122,175.5) -- (122,181.5);

\draw[wire] (229,116) -- (242,116) -- (242,189.5) -- (232,189.5);
\draw[arr] (232,189.5) -- (226,189.5);
\node[fill=white,inner sep=1pt] at (242,160) {$h$};

\foreach \x/\lab/\col in {43/Actions/red,122/States/blue,201/Proxy/magenta} {
  \node[module,fill=\col!18,minimum width=65pt,minimum height=16pt]
    at (\x,211.5) {Future \lab};
  \draw[arr] (\x,197.5) -- (\x,203.5);
}

\node[font=\fontsize{7}{8}\selectfont] at (122,228)
  {Stage II: fine-tune backbone and action expert};
\end{tikzpicture}}
    \caption{\textbf{\vtamc architecture.} Stage I fine-tunes the video backbone to predict future latents for two RGB streams and one tactile stream, with in-view attention shared across streams. Stage II fine-tunes it end-to-end with a parallel expert that jointly denoises actions, proprioceptive states, and a flow-force proxy.}
    \label{fig:vtam_architecture}
    \vspace{-0.4cm}
\end{figure}

\noindent\textbf{Stage I objective.}
Let $x_0$ be the clean latent sequence of future RGB and tactile observations, and let $c_{\mathrm{obs}}$ denote the conditioning observations. We use flow matching~\cite{lipman2022flow} with a noise sample $\epsilon$ and flow time $\tau\in[0,1]$:
\begin{equation}
x_\tau=(1-\tau)x_0+\tau\epsilon,\qquad v^*=\epsilon-x_0.
\end{equation}
The predictive objective is
\begin{equation}
\mathcal L_{\mathrm{stage1}}=
\mathbb E\!\left[\left\|v_\theta(x_\tau,\tau\mid c_{\mathrm{obs}})-v^*\right\|^2\right].
\end{equation}
The loss excludes conditioning frames and adapts the representation through future sensory prediction. Throughout, $k$ denotes observation time and $\tau$ flow time.
Initializing Stage II from this tactile world model lets sensory adaptation precede policy optimization, so the auxiliary loss shapes both the backbone features and the expert's use of them.

\subsection{Flow-Force Proxy Supervision}
We derive a \emph{flow-force proxy}, a compact force-like target computed from tactile optical flow, and the action expert predicts it as an auxiliary Stage~II target. Because the expert cross-attends to both visual and tactile backbone tokens, this supervision encourages it to draw on tactile tokens when contact occurs rather than collapsing onto visual features. Direct force supervision is impractical for a parallel gripper: opposing fingertip forces cancel at a wrist force/torque sensor, in-sensor force sensing adds cost and bulk, and Sharpa trains a separate force regressor~\cite{su2026tacmap}. Our proxy needs no force sensor, calibration, or learned estimator, and adds negligible computation; removing its supervision sharply lowers success (Section~\ref{sec:ablations}).\looseness-1

Following the use of tactile deformation to characterize contact~\cite{yuan2015shear}, we compute dense optical flow $u_k=(u_{x,k},u_{y,k})$ between a no-contact reference $I_{\mathrm{ref}}$ and the current tactile image $I_k^3$. 
Spatial averaging yields a three-dimensional descriptor:
\begin{equation}
\begin{aligned}
f_{x,k}&=\mathbb E_{p}[u_{x,k}(p)],\quad
f_{y,k}=\mathbb E_{p}[u_{y,k}(p)],\\
f_{z,k}&=\mathbb E_{p}[\nabla\!\cdot u_k(p)],\quad
F_k^v=[f_{x,k},f_{y,k},f_{z,k}]^\top.
\end{aligned}
\end{equation}
Here $p$ indexes image locations. The first two components summarize tangential image displacement, and the third summarizes expansion or contraction of the deformation field. We call $F_k^v$ the flow-force proxy.

\subsection{Conditional Action--State--Proxy Learning}
In Stage II, a 28-layer action expert attends to the corresponding hidden features $h$ of the video backbone, which is initialized from Stage I and fine-tuned jointly with the expert. Each expert layer contains self-attention over the joint prediction tokens, cross-attention to the backbone features, and a feed-forward module; adaptive layer normalization conditions the layer on flow time.

The clean joint target is $y_0=[a;f;s]$, concatenated over the prediction horizon, where each timestep contains a seven-dimensional action command, a three-dimensional flow-force proxy, and a 16-dimensional proprioceptive state. The current-state conditioning token is $c_k=[\mathbf 0_{10};s_k]$, with zero-padded action and proxy entries. The action, state, and proxy outputs are components of one joint denoising target. Their losses supervise corresponding slices of the predicted velocity field, so the proxy is learned through the same expert that generates actions. For a noisy joint target $y_\tau$, the component objectives are
\begin{figure*}[!t]
    \centering
\includegraphics[width=0.95\linewidth]{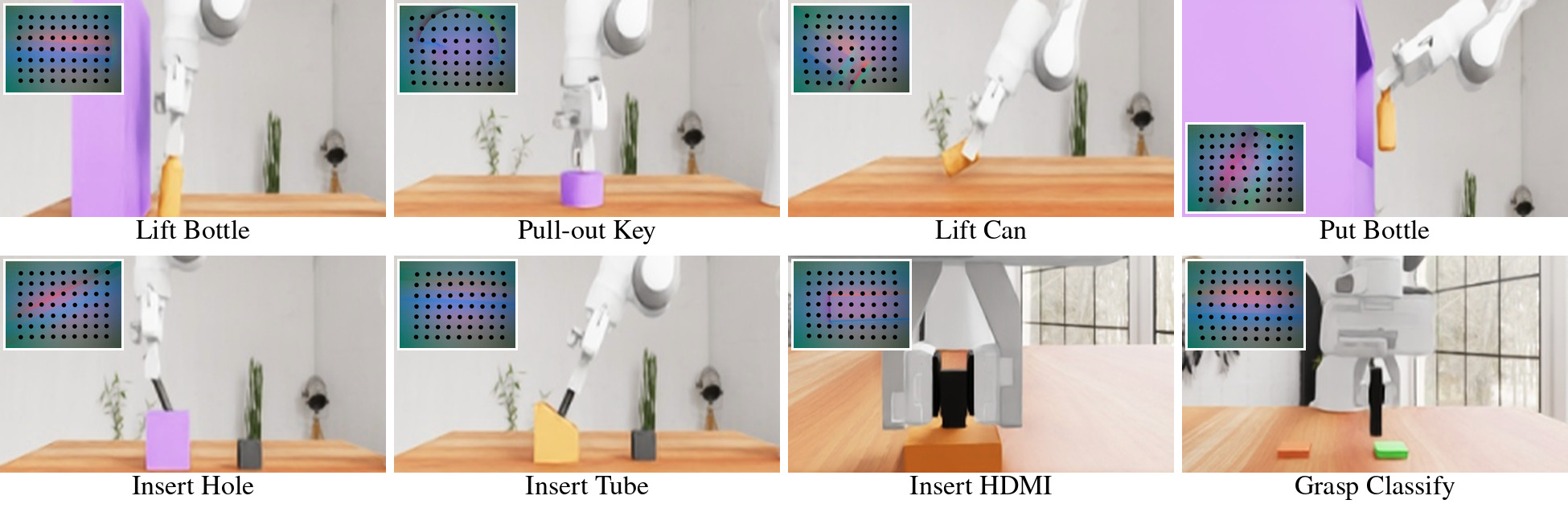}
    \caption{\textbf{The eight UniVTAC simulation tasks.} Each tile shows a head-camera frame from a demonstration near the key contact moment, with the GelSight tactile image inset.}
    \label{fig:univtac_tasks}
    \vspace{-0.2cm}
\end{figure*}
\begin{equation}
\label{eq: action}
\begin{aligned}
\mathcal L_{\mathrm{action}}&=\mathbb E\!\left[\left\|v_\phi^a(y_\tau,\tau\mid h,c_k)-v^{*a}\right\|^2\right],\\
\mathcal L_{\mathrm{state}}&=\mathbb E\!\left[\left\|v_\phi^s(y_\tau,\tau\mid h,c_k)-v^{*s}\right\|^2\right].
\end{aligned}
\end{equation}
The flow-force proxy loss is computed in the same joint denoising process:
\begin{equation}
\label{eq:force}
\mathcal L_{\mathrm{force}}=\mathbb E\!\left[\left\|v_\phi^f(y_\tau,\tau\mid h,c_k)-v^{*f}\right\|^2\right].
\end{equation}
The complete objective is
\begin{equation}
\mathcal L_{\mathrm{stage2}}=\mathcal L_{\mathrm{action}}+
\lambda_1\mathcal L_{\mathrm{state}}+\lambda_2\mathcal L_{\mathrm{force}},
\end{equation}
with $\lambda_1=\lambda_2=1$ in the reported experiments. The velocity targets use the corresponding normalized data components and noise samples. State prediction supplies a dynamics-related target, while proxy loss supervises tactile-deformation information in the joint output.

At deployment, iterative video denoising is not run; it serves only as a self-supervised training objective. A single backbone forward pass, with the future latent slots set to noise at $\tau=1$, yields per-layer features that are cached. The expert then integrates its flow over a few Euler steps, attending to these features and the current state, to generate an action chunk.\looseness-1

\begin{table*}[t]

\centering
\footnotesize
\setlength{\tabcolsep}{8pt}
\newcommand{\hd}[2]{\textbf{\begin{tabular}[c]{@{}c@{}}#1\\#2\end{tabular}}}
\newcommand{\grp}[2]{\begin{tabular}[c]{@{}c@{}}#1\\#2\end{tabular}}
\newcommand{\g}[1]{\textcolor{gray}{#1}}
\caption{Success rates (\%) on the eight UniVTAC tasks (H/W: head/wrist RGB views). Bold and underline mark the best and second best without tactile pretraining.}
\label{tab:univtac_sim}
\begin{tabular}{c l c c c c c c c c c}
    \toprule
    \rowcolor[HTML]{EFEFEF}
     & \textbf{Model} & \hd{Lift}{Bottle} & \hd{Pull-out}{Key} & \hd{Lift}{Can} & \hd{Put}{Bottle} & \hd{Insert}{Hole} & \hd{Insert}{Tube} & \hd{Insert}{HDMI} & \hd{Grasp}{Classify} & \textbf{Avg.} \\
    \rowcolor[HTML]{EFEFEF}
     & \textit{RGB views} & H & H & H+W & H & H & H+W & H & H & \\
    \midrule
    \multirow{3}{*}{\grp{Vision-only}{Policies}}
    & $\pi_{0.5}$~\cite{intelligence2025pi_} & \textbf{100} & 35 & 6 & 34 & 25 & 74 & 8 & 49 & 41.4 \\
    & LingBot-VA~\cite{li2026lingbotva} & 0 & 58 & 0 & 0 & 42 & 96 & \underline{38} & 17 & 31.4 \\
    & Fast-WAM~\cite{yuan2026fastwam} & 21 & \underline{73} & 0 & 35 & \underline{66} & \textbf{98} & 19 & 72 & \underline{48.0} \\
    \midrule
    \multirow{3}{*}{\begin{tabular}[c]{@{}c@{}}Visuo-tactile\\w/o Tactile\\Pretraining\end{tabular}}
    & ACT + UniVTAC~\cite{chen2026univtac} & 71 & 46 & \underline{29} & 31 & 24 & 56 & 28 & \underline{99} & \underline{48.0} \\
    & Tactile-VLA~\cite{huang2025tactile} & \underline{97} & 32 & 15 & 10 & 41 & 56 & 12 & \textbf{100} & 45.4 \\
    & Tactile-WAM~\cite{wu2026tactile} & 5 & 20 & 10 & \underline{45} & 20 & 85 & 0 & 55 & 30.0 \\
    \midrule
    \multirow{4}{*}{\g{\begin{tabular}[c]{@{}c@{}}Visuo-tactile\\w/ Tactile\\Pretraining\end{tabular}}}
    & \g{FTP-1~\cite{yuan2026ftp}} & \g{97} & \g{48} & \g{65} & \g{47} & \g{64} & \g{79} & \g{18} & \g{100} & \g{64.8} \\
    & \g{N$_0$-VTLA~\cite{neoteai2026vtla}} & \g{99} & \g{99} & \g{88} & \g{60} & \g{95} & \g{99} & \g{25} & \g{100} & \g{83.1} \\
    & \g{N$_0$-TWAM (20\%)~\cite{neoteai2026twam}} & \g{51} & \g{25} & \g{80} & \g{46} & \g{78} & \g{96} & \g{52} & \g{95} & \g{65.4} \\
    & \g{N$_0$-TWAM~\cite{neoteai2026twam}} & \g{58} & \g{79} & \g{93} & \g{87} & \g{99} & \g{98} & \g{68} & \g{94} & \g{84.5} \\
    \midrule
    \rowcolor{IllinoisOrange!10}
     & \textbf{\vtamc (Ours)} & 91 & \textbf{92} & \textbf{75} & \textbf{100} & \textbf{67} & \underline{97} & \textbf{45} & \textbf{100} & \textbf{83.4} \\
    \bottomrule
\end{tabular}
\end{table*}

\section{Experiments}
\label{sec:experiments}

We first evaluate on UniVTAC~\cite{chen2026univtac}, a visuo-tactile simulation benchmark, on eight contact-rich tasks (Fig.~\ref{fig:univtac_tasks}, Table~\ref{tab:univtac_sim}). We evaluate 100 seeds per task with the official 300-step budget (600 steps for Insert HDMI). \vtam is fine-tuned only on the official UniVTAC demonstrations, without tactile or action pretraining. Baseline numbers are taken from prior work~\cite{yuan2026ftp,wu2026tactile,chen2026univtac,neoteai2026vtla,neoteai2026twam} where available (Tactile-WAM reports 20 trials per task) and evaluated by us otherwise; on Put Bottle, \vtam uses a 24-step execution horizon.\looseness-1

For real-world experiments, we evaluate \vtam on a six-DoF xArm6 with a parallel gripper, one finger-mounted GelSight Mini sensor, and two Intel RealSense D455 cameras. The policy uses the RGB streams together with tactile images and robot state. 
Each model is evaluated on 80 real-world trials: 20 consecutive chip trials, 20 peeling motions, and 20 wiping trials on each of a flat and a tilted board:\looseness-1

\noindent \textbf{Chip Pick-and-Place} (Chip) requires grasping a potato chip from a newly randomized starting position in each trial and transferring it to a plate. Missing the grasp, breaking the chip, or dropping it during transport counts as failure.\looseness-1

\noindent \textbf{Cucumber Peeling} (Peel) comprises 20 consecutive motions at a fixed cutting position. Success requires a peel strip longer than 10\,cm. The surface height decreases during the sequence, so these motions share an evolving object rather than independently reset geometry.

\noindent \textbf{Whiteboard Wiping} (Wipe) uses a rigid eraser on a flat or tilted board. A black stain is redrawn randomly at the start of each trial. The policy is allowed at most five wiping motions; success requires complete stain removal without disturbing the board or its supports.
Manual teleoperation provides 100 chip pick-and-place, 61 cucumber peeling, and 105 whiteboard wiping trajectories with synchronized observations. We train a separate task-specific policy for each setting.

\noindent \textbf{Hanoi Disk Insertion} (Hanoi) tests sensor transfer, using a seven-DoF Flexiv Rizon 4 arm, a gripper-mounted Xense Photon sensor instead of GelSight Mini, and one wrist camera. The robot places one disk onto the tower peg; success requires it to be seated. We use 100 demonstrations and 20 trials (Table~\ref{tab:hanoi_xense}).

\subsection{Training and Baselines}\label{sec:implementation}\label{app:details}
\vtam is initialized from GE-base~\cite{liao2025genie} and trained on four A100 GPUs with AdamW~\cite{loshchilov2017decoupled}, batch size 16 per GPU, and 1,000 warmup steps. Stage~I trains the visual and tactile future-latent objective for 50,000 steps at a learning rate of $3\times10^{-4}$; Stage~II fine-tunes the backbone jointly with the parallel 28-layer action expert for 20,000 steps at $5\times10^{-5}$, with the VAE frozen. All tasks use $192\times256$ images, nine-frame video chunks, and 54-step action chunks.

We compare against GE-Act~\cite{liao2025genie}, vision-only $\pi_{0.5}$~\cite{intelligence2025pi_}, $\pi_{0.5}$ with GelSight supplied as an additional image view, RDP~\cite{xue2025reactive}, and Tactile-VLA~\cite{huang2025tactile}. GE-Act is the vision-only counterpart of our model: it uses the same GE-base initialization and 50,000/20,000-step schedule, with a randomly initialized action expert in Stage~II and all other settings matched. Both $\pi_{0.5}$ variants fine-tune the official pretrained checkpoint for 10,000 steps on the same hardware. On the real robot, all methods are trained on the same demonstrations and evaluated with the same protocol.\looseness-1

\noindent\textbf{Inference and execution rates.}
Evaluation uses 1\,Hz policy inference, with data collection and low-level action execution at 30\,Hz. The policy generates action chunks from each observation update, which sets the tactile feedback rate to 1\,Hz.\looseness-1

\section{Results and Analysis}
\label{sec:results}
\subsection{Simulation Results}
\label{sec:sim_results}
Table~\ref{tab:univtac_sim} reports success rates on the eight UniVTAC tasks~\cite{chen2026univtac}. Among methods without large-scale visuo-tactile pretraining, \vtam outperforms every baseline on five of the eight tasks, ties Tactile-VLA on Grasp Classify, and falls short of the best baseline on Lift Bottle (91\% vs.\ 100\% ($\pi_{0.5}$)) and Fast-WAM on Insert Tube (97\% vs.\ 98\%), reaching an 83.4\% average versus 48.0\% for the strongest such baselines (Fast-WAM and ACT + UniVTAC). Without tactile or action pretraining, it also exceeds the tactile-pretrained FTP-1 (64.8\%) and N$_0$-TWAM with 20\% of its pretraining data (65.4\%), and is comparable to N$_0$-VTLA (83.1\%) and N$_0$-TWAM (84.5\%), both pretrained on NeoData, over 30K hours of largely visuo-tactile robot data, whereas \vtam uses only the provided task demonstrations.

\subsection{Real-Robot Task Performance}
\vtam outperforms every baseline on every real-robot task. Table~\ref{tab:overall_performance} reports 90\%, 85\%, and 95\% on chip pick-and-place, peeling, and wiping. On the chip task, the vision-only $\pi_{0.5}$ baseline achieves 10\%, an 80-percentage-point gap. On peeling, Tactile-VLA is the only baseline with nonzero success (20\%). On wiping, RDP is the strongest evaluated baseline at 50\%, compared with 95\% for \vtam.
\begin{table}[t]

\centering
\small
\caption{\textbf{Task success:} 20 chip trials, 20 peeling motions, and 40 wiping trials (flat/tilted). Avg.\ is the unweighted mean of the three task rates.}
\label{tab:overall_performance}

\begin{tabular}{l c c c c}
    \toprule
    \rowcolor[HTML]{EFEFEF}
    \textbf{Model} & \textbf{Chip} & \textbf{Peel} & \textbf{Wipe} & \textbf{Avg.} \\
    \midrule
    Genie Envisioner & 0\% & 0\% & 2.5\% & 0.8\% \\
    \rowcolor[HTML]{F9F9F9}
    $\pi_{0.5}$ (Vision) & 10\% & 0\% & 0\% & 3.3\% \\
    $\pi_{0.5}$ + Tactile & 5\% & 0\% & 0\% & 1.7\% \\
    RDP & 0\% & 0\% & 50\% & 16.7\% \\
    Tactile-VLA~\cite{huang2025tactile} & 0\% & 20\% & 10\% & 10.0\% \\
    \midrule
    \rowcolor{IllinoisOrange!10}
    \textbf{\vtamc (Ours)} & \textbf{90\%} & \textbf{85\%} & \textbf{95\%} & \textbf{90.0\%} \\
    \bottomrule
\end{tabular}
\vspace{-0.2cm}
\end{table}

\begin{figure*}[!t]
\centering

\newcommand{\comparisonpanel}[2]{%
\begin{minipage}[t]{0.32\textwidth}
\centering
\vspace{0pt}
\begin{tikzpicture}
\node[inner sep=0pt] (img)
 {\includegraphics[width=0.95\linewidth]
 {#1}};

\node[anchor=east,
      font=\fontsize{6}{7}\selectfont\bfseries,inner sep=2pt]
 at ($(img.north west)!0.25!(img.south west)$)
 {\rotatebox{90}{Baseline}};

\node[anchor=east,
      font=\fontsize{6}{7}\selectfont\bfseries,inner sep=2pt]
 at ($(img.north west)!0.75!(img.south west)$)
 {\rotatebox{90}{\vtam (Ours)}};

\node[anchor=north,
      font=\fontsize{7}{8}\selectfont,inner sep=2pt]
 at (img.south) {#2};
\end{tikzpicture}
\end{minipage}%
}

\comparisonpanel{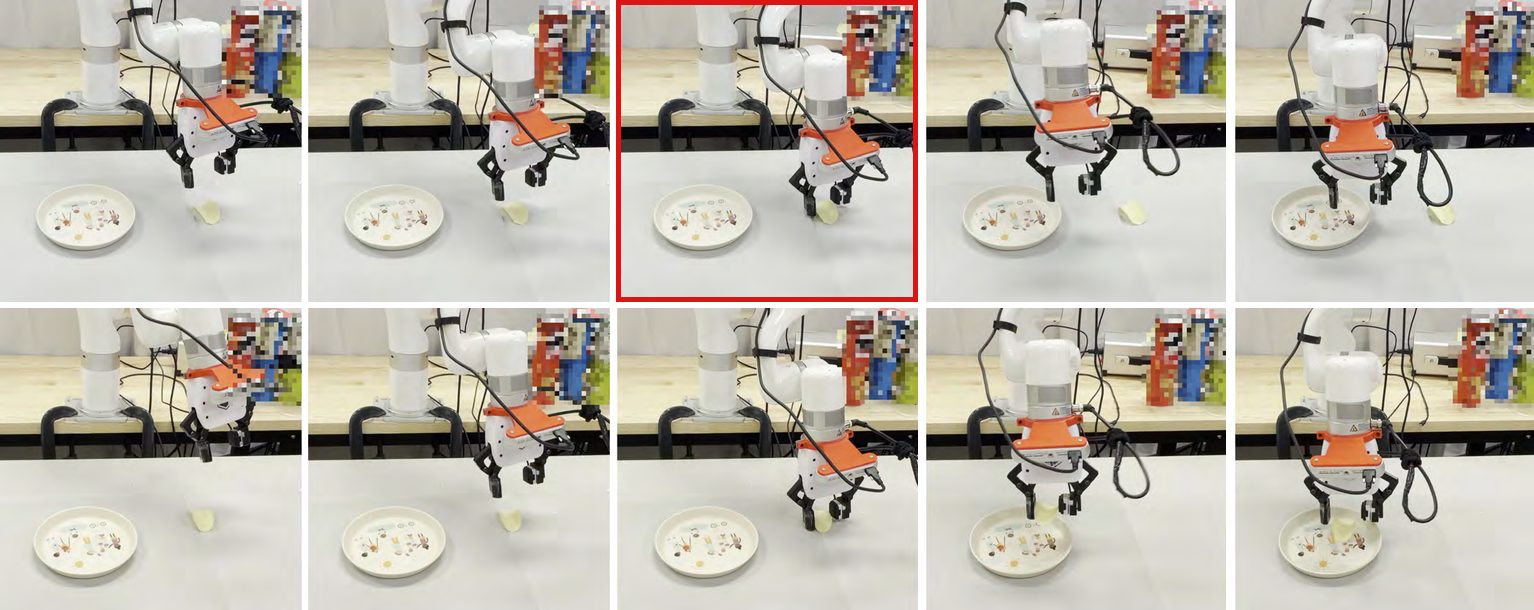}{(a) Chip pick-and-place}\hfill
\comparisonpanel{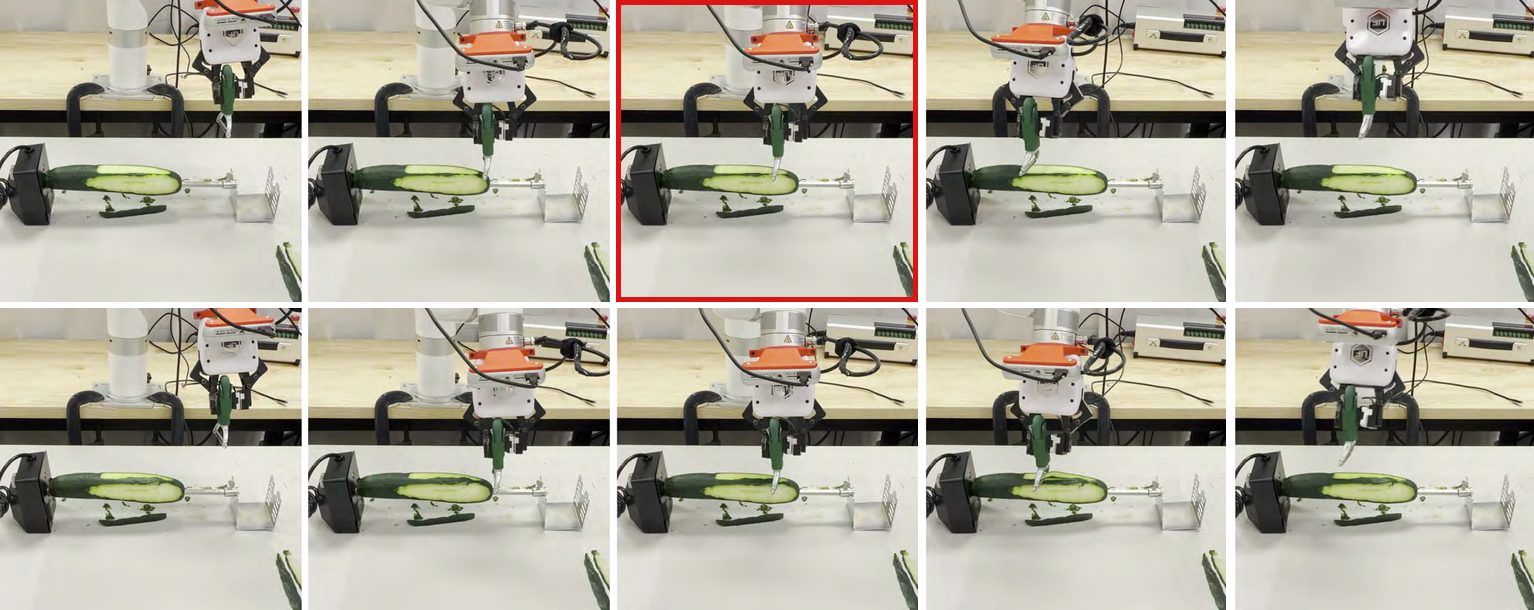}{(b) Peeling}\hfill
\comparisonpanel{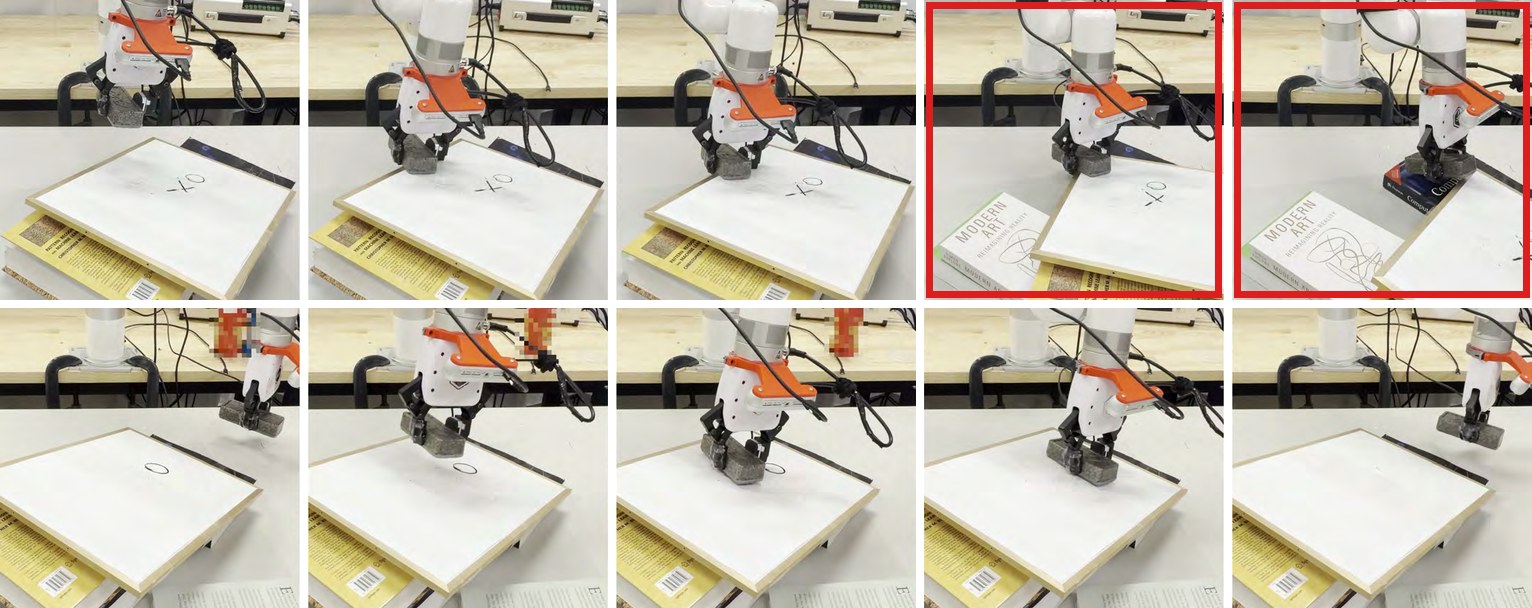}{(c) Wiping}

\setcounter{figure}{5}\caption{\textbf{Real-robot comparisons.}
The baseline (a) misses the chip, (b) loses contact during
peeling, and (c) pushes the board off its supports (red boxes).
\vtam successfully places the chip and maintains tool contact
during peeling and wiping.\looseness-1}
\label{fig:qualitative}
\end{figure*}

\begin{table}[t!]
\centering
\footnotesize
\setlength{\tabcolsep}{2.5pt}
\caption{\textbf{Wiping tactile summaries.} SCR: stable contact ratio ($\uparrow$); S$_q$: signal quantiles ($\downarrow$); TV/frame: temporal variation ($\downarrow$).}
\label{tab:tactile_wipe_quality}
\begin{tabular}{lccccc}
\toprule
 \rowcolor[HTML]{EFEFEF}  \textbf{Model}  & \textbf{SCR} $\uparrow$ & \textbf{S}$_{q50}\downarrow$ & \textbf{S}$_{q95}\downarrow$ & \textbf{S}$_{q99}\downarrow$ & \textbf{TV/frame} $\downarrow$ \\
\midrule
RDP  & 0.750 & 9.12 & 12.98 & 13.71 & 0.153 \\
$\pi_{0.5}$   & 0.317 & 12.08 & 19.59 & 21.33 & 0.540 \\
\midrule
 \rowcolor{IllinoisOrange!10} \textbf{\vtamc (Ours)} & \textbf{0.945} & \textbf{7.45} & \textbf{8.53} & \textbf{9.58} & \textbf{0.144} \\
\bottomrule
\end{tabular}
\vspace{-0.2cm}
\end{table}

Table~\ref{tab:tactile_wipe_quality} reports sensor-derived wiping summaries. Since an eraser that loses contact also gives a weak signal, we read the signal quantiles together with SCR and success. \vtam has higher SCR than RDP (0.945 versus 0.750), lower quantiles, and slightly lower TV/frame (0.144 versus 0.153), consistent with more sustained contact and less variable tactile response.

On Hanoi Disk Insertion with an Xense Photon sensor instead of GelSight Mini (Table~\ref{tab:hanoi_xense}), \vtam succeeds in 15 of 20 trials (75\%), compared with 55\% for the tactile-pretrained FTP-1, 25\% for vision-only $\pi_{0.5}$, 5\% for Genie Envisioner, and 0\% for Tactile-VLA. The same architecture thus works with a different sensor. Fig.~\ref{fig:qualitative_hanoi} shows a representative rollout.

\subsection{Qualitative Contact Behavior}\label{sec:qualitative_behavior}\label{app:examples}\label{app:ablation}
Figure~\ref{fig:qualitative} shows representative task rollouts. GE-Act and the $\pi_{0.5}$ variants can close above the chip and continue toward the target plate after a failed grasp. \vtam establishes contact before transport and maintains a stable grasp.
During cucumber peeling, GE-Act and the $\pi_{0.5}$ variants move along the cucumber but can lose contact with its surface. \vtam maintains contact and completes repeated peeling motions as the cucumber height decreases.
During whiteboard wiping, the baseline displaces the board supports, whereas \vtam keeps the eraser in contact and removes the stain without moving the board.

\subsection{Component Ablation and Tactile Utilization}
\label{sec:ablations}

\begin{figure}[t!]
\centering
\begin{tikzpicture}
\node[inner sep=0pt] (img) {\includegraphics[width=0.85\linewidth]{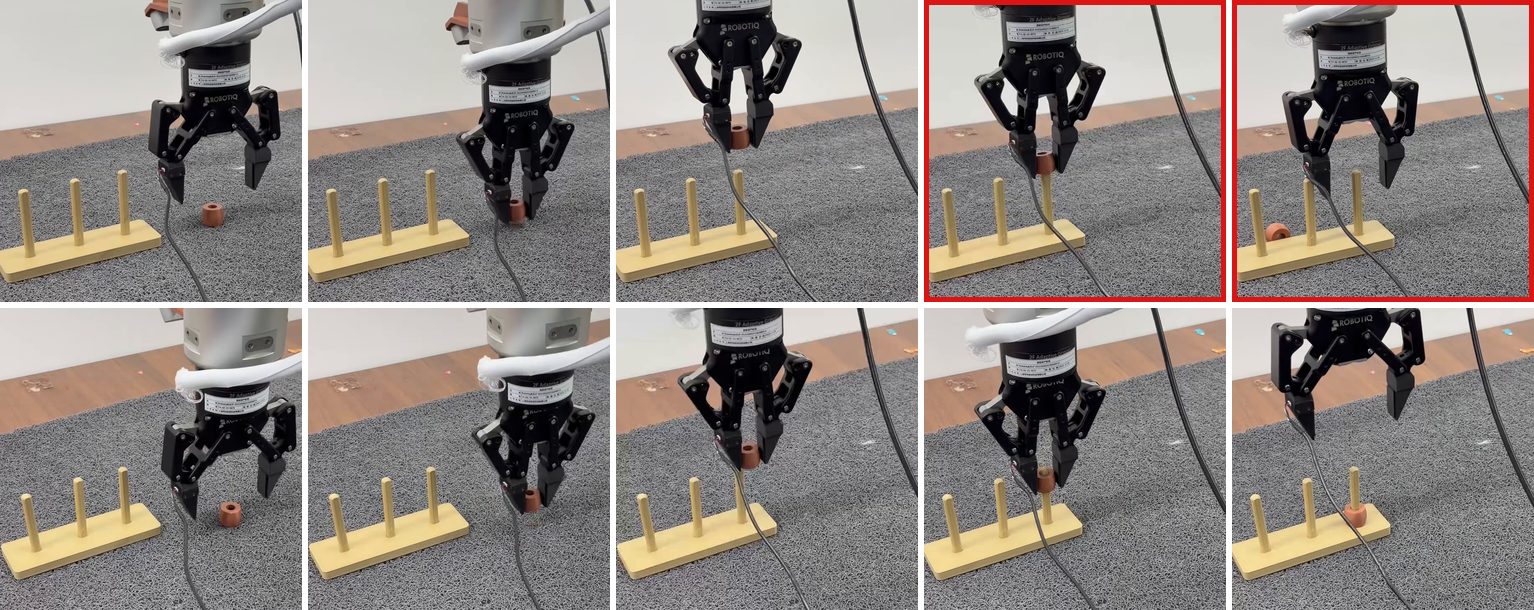}};
\node[anchor=east,
      font=\fontsize{6}{7}\selectfont\bfseries,inner sep=2pt]
 at ($(img.north west)!0.25!(img.south west)$)
 {\rotatebox{90}{Baseline}};

\node[anchor=east,
      font=\fontsize{6}{7}\selectfont\bfseries,inner sep=2pt]
 at ($(img.north west)!0.75!(img.south west)$)
 {\rotatebox{90}{\vtam (Ours)}};
\end{tikzpicture}
\setcounter{figure}{4}\caption{\textbf{Hanoi Disk Insertion.} The baseline presses the disk against the peg tip and releases it off the peg (red boxes); \vtam seats the disk on the peg.}
\label{fig:qualitative_hanoi}
\vspace{-0.2cm}
\end{figure}
\begin{table}[t!]

\centering
\small
\caption{\textbf{Hanoi Disk Insertion with a Xense Photon sensor} (sensor generalization).}
\label{tab:hanoi_xense}

\begin{tabular}{l c c}
    \toprule
    \rowcolor[HTML]{EFEFEF}
    \textbf{Model} & \textbf{Successes} & \textbf{Success Rate} \\
    \midrule
    Genie Envisioner & 1/20 & 5\% \\
    \rowcolor[HTML]{F9F9F9}
    $\pi_{0.5}$ (Vision) & 5/20 & 25\% \\
    Tactile-VLA~\cite{huang2025tactile} & 0/20 & 0\% \\
    \rowcolor[HTML]{F9F9F9}
    FTP-1~\cite{yuan2026ftp} & 11/20 & 55\% \\
    \midrule
    \rowcolor{IllinoisOrange!10}
    \textbf{\vtamc (Ours)} & \textbf{15/20} & \textbf{75\%} \\
    \bottomrule
\end{tabular}
\vspace{-0.2cm}
\end{table}

\noindent\textbf{Real-robot component ablations.}
Table~\ref{tab:ablation} reports real-robot component ablations.
Feeding frozen-VAE tactile features directly to the action expert, with the flow-force proxy supervision kept, yields no successes on any of the three tasks. Removing this target from \vtam reduces success from 90\% to 10\% on chip pick-and-place, from 95\% to 10\% on wiping, and from 75\% to 55\% on Hanoi Disk Insertion. These results support the complementary roles of tactile world modeling and contact-related supervision: the backbone must learn useful tactile representations, while the auxiliary loss encourages their use during action learning.\looseness-1

\begin{table}[t]
\centering
\footnotesize
\caption{Real-robot component ablation: success rates of \vtam and two variants on Chip Pick-and-Place, Whiteboard Wiping, and Hanoi Disk Insertion (Xense Photon).}
\label{tab:ablation}
\setlength{\tabcolsep}{3pt}
\begin{tabular}{l c c c c}
    \toprule
    \rowcolor[HTML]{EFEFEF}
    \textbf{Model Variant} & \textbf{Tactile Features} & \textbf{Chip} & \textbf{Wipe} & \textbf{Hanoi} \\
    \midrule
    Tactile VAE $\rightarrow$ expert & Frozen VAE & 0\% & 0\% & 0\% \\
    \rowcolor[HTML]{F9F9F9}
    w/o flow-force proxy & World model & 10\% & 10\% & 55\% \\
    \rowcolor{IllinoisOrange!10}
    \textbf{\vtamc (Ours)} & \textbf{World model} & \textbf{90\%} & \textbf{95\%} & \textbf{75\%} \\
    \bottomrule
\end{tabular}
\vspace{-0.2cm}
\end{table}

\begin{table}[t]
\centering
\footnotesize
\newcommand{\hd}[2]{\textbf{\begin{tabular}[c]{@{}c@{}}#1\\#2\end{tabular}}}
\caption{\textbf{Effect of Stage I visuo-tactile world-model fine-tuning} on UniVTAC (success \%).}
\label{tab:ablation_stage1}
\setlength{\tabcolsep}{1.6pt}
\resizebox{\linewidth}{!}{%
\begin{tabular}{l c c c c c c c c c}
    \toprule
    \rowcolor[HTML]{EFEFEF}
    \textbf{Variant} & \hd{Lift}{Bottle} & \hd{Pull-out}{Key} & \hd{Lift}{Can} & \hd{Put}{Bottle} & \hd{Insert}{Hole} & \hd{Insert}{Tube} & \hd{Insert}{HDMI} & \hd{Grasp}{Classify} & \textbf{Avg.} \\
    \midrule
    \vtamc w/o Stage I & 27 & 72 & 49 & 54 & \textbf{95} & 91 & 38 & \textbf{100} & 65.8 \\
    \rowcolor{IllinoisOrange!10}
    \textbf{\vtamc (full)} & \textbf{91} & \textbf{92} & \textbf{75} & \textbf{100} & 67 & \textbf{97} & \textbf{45} & \textbf{100} & \textbf{83.4} \\
    \bottomrule
\end{tabular}}
\vspace{-0.2cm}
\end{table}

\begin{table}[t]
\centering
\footnotesize
\newcommand{\hd}[2]{\textbf{\begin{tabular}[c]{@{}c@{}}#1\\#2\end{tabular}}}
\caption{\textbf{Backbone adaptation} on UniVTAC (success \%). Frozen trains only the action expert in Stage II.}
\label{tab:ablation_frozen}
\setlength{\tabcolsep}{3pt}
\begin{tabular}{c c c c c}
    \toprule
    \rowcolor[HTML]{EFEFEF}
    \textbf{Stage I} & \textbf{Stage II backbone} & \hd{Lift}{Bottle} & \hd{Put}{Bottle} & \hd{Insert}{Hole} \\
    \midrule
    $\times$ & frozen & 4 & 6 & 53 \\
    $\times$ & fine-tuned & 27 & 54 & \textbf{95} \\
    $\checkmark$ & frozen & 18 & 10 & 0 \\
    \rowcolor{IllinoisOrange!10}
    $\checkmark$ & fine-tuned (\vtamc) & \textbf{91} & \textbf{100} & 67 \\
    \bottomrule
\end{tabular}
\end{table}

\noindent\textbf{Stage I world-model fine-tuning.} Skipping Stage I and training Stage II directly from the pretrained video model lowers success on most UniVTAC tasks (Table~\ref{tab:ablation_stage1}): Lift Bottle drops from 91\% to 27\%, Put Bottle from 100\% to 54\%, Lift Can from 75\% to 49\%, and Pull-out Key from 92\% to 72\%, and the eight-task average falls from 83.4\% to 65.8\%. Stage I world-model fine-tuning thus improves the visuo-tactile representation used by the action expert in most cases, even though Stage II further fine-tunes the backbone jointly with the action expert. Grasp Classify saturates with or without it, and Insert Hole is an exception where \vtam w/o Stage I succeeds more often (95\% vs.\ 67\%).

\noindent\textbf{Backbone adaptation.} Table~\ref{tab:ablation_frozen} trains only the action expert with the backbone frozen. Without Stage I, frozen video features nearly fail on Lift Bottle (4\%) and Put Bottle (6\%), well below end-to-end fine-tuning (27\% and 54\%) and full \vtam (91\% and 100\%), so the backbone itself must adapt to touch rather than serve as a fixed feature extractor. Freezing the Stage I backbone yields only 18\% and 10\% success on Lift Bottle and Put Bottle, respectively, substantially below the results achieved with end-to-end Stage II fine-tuning. These results support the complementary roles of Stage I adaptation and end-to-end Stage II fine-tuning. Insert Hole again differs: the frozen pretrained video model still reaches 53\%, whereas the frozen Stage I backbone reaches 0\%, and end-to-end fine-tuning recovers 67\%. On this task Stage I moves the representation away from what the policy needs, which we leave to future study.

\begin{figure}[!t]
\centering
\subfloat[w/o proxy supervision]{\includegraphics[width=0.49\linewidth,trim=0 0 675bp 0,clip]{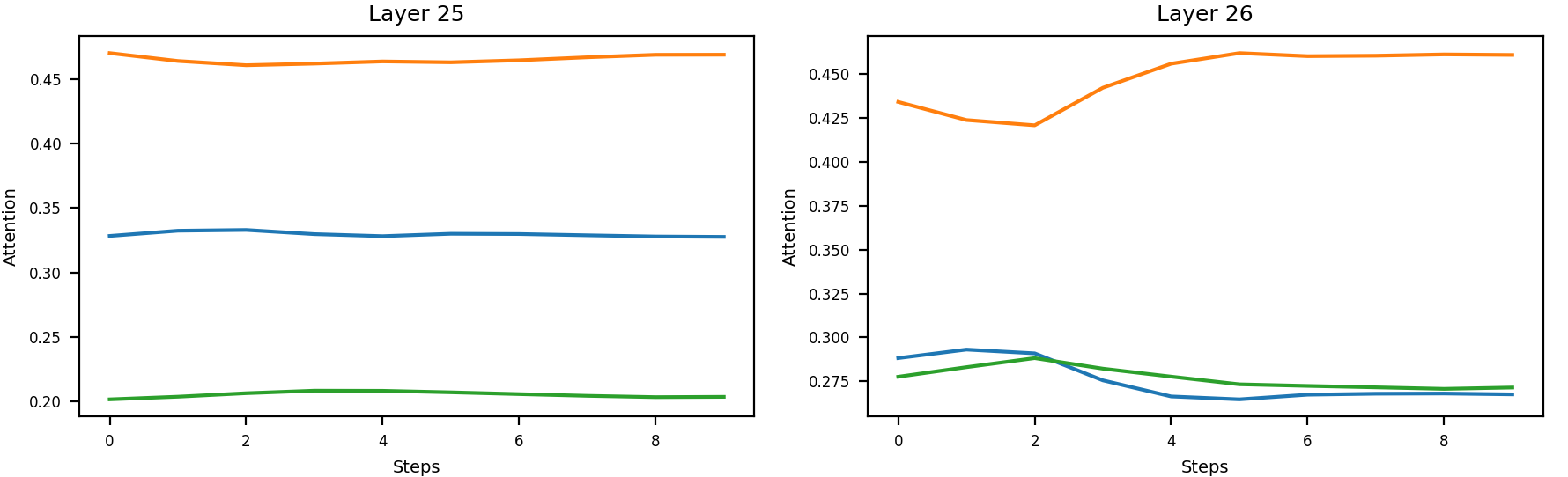}}\hfill
\subfloat[with proxy supervision]{\includegraphics[width=0.49\linewidth,trim=0 0 675bp 0,clip]{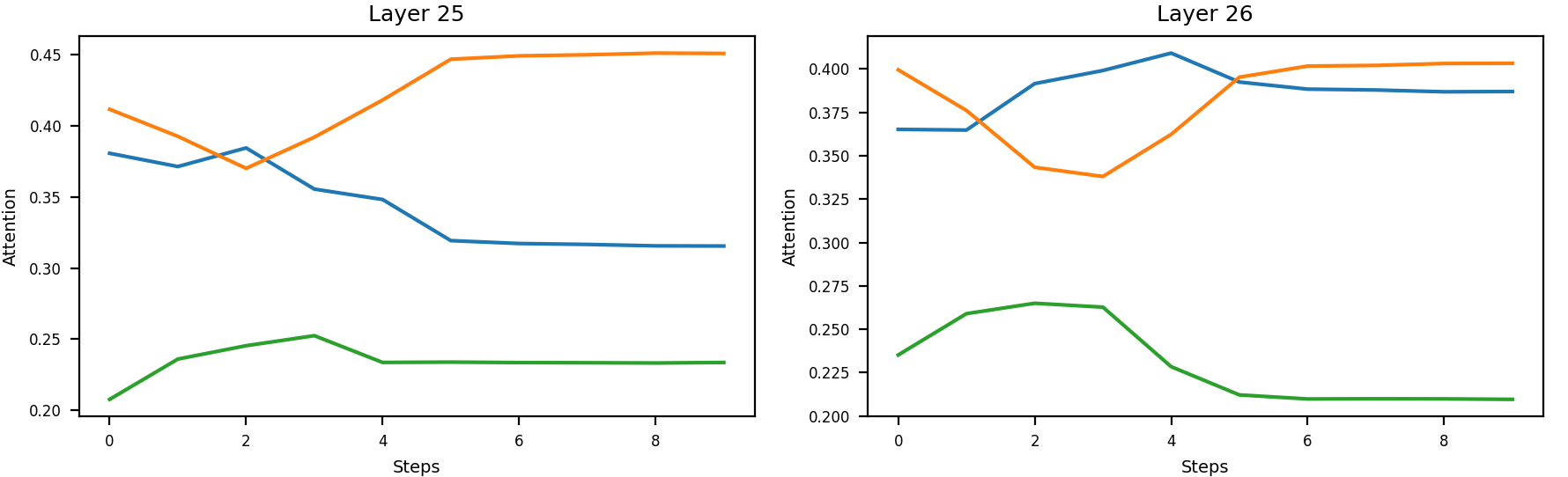}}
\setcounter{figure}{7}\caption{\textbf{Cross-attention share of the action expert (layer 25)} over side-camera (blue), front-camera (orange), and GelSight (green) views on a real chip pick-and-place episode, (a) w/o and (b) with the flow-force proxy supervision. Steps index open-loop action chunks (1.8\,s each).}
\label{fig:attention_modulation}
\vspace{-0.3cm}
\end{figure}

\noindent\textbf{Flow-force proxy supervision and modality collapse.} Removing only the flow-force proxy supervision but keeping the tactile world-model features reduces the success rate to 10\% on chip and wiping and 55\% on Hanoi Disk Insertion. Fig.~\ref{fig:attention_modulation} shows this modality collapse: at layer 25, with the proxy supervision, the tactile share rises during contact (mean tactile share in contact vs.\ outside: 0.26 vs.\ 0.22), whereas without it the share stays nearly flat (0.21 vs.\ 0.20). The proxy thus keeps tactile attention responsive to contact.

\subsection{Sensory Prediction}\label{sec:sensory_prediction}\label{app:videoexamples}


\begin{figure}[!t]
\centering
\begin{tikzpicture}[
  panel/.style={
    anchor=north,
    inner sep=0pt,
    outer sep=0pt
  },
  lab/.style={
    font=\fontsize{6}{7}\selectfont\bfseries,
    inner sep=1pt
  },
  heading/.style={
    font=\fontsize{7}{8}\selectfont\bfseries,
    inner sep=1pt
  }
]

\node[panel] (wr) at (0,0)
 {\includegraphics[width=\dimexpr\linewidth-30pt\relax]
 {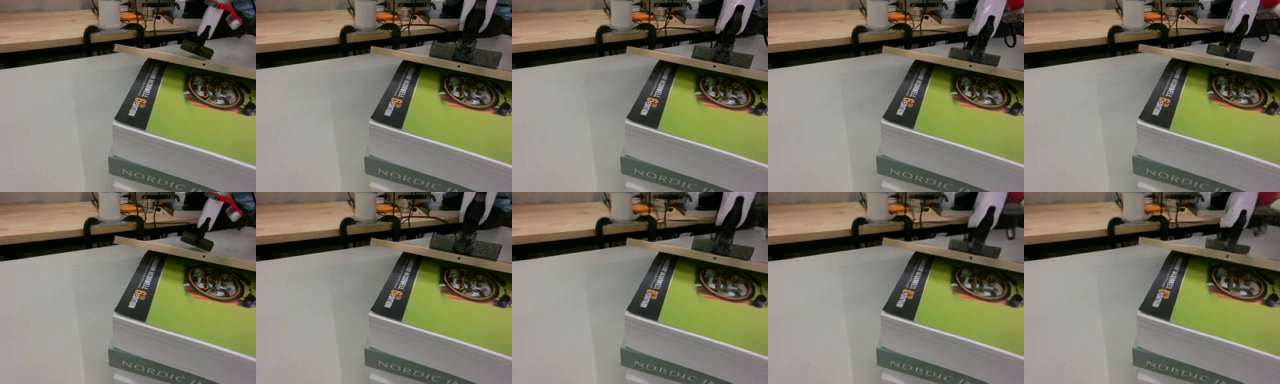}};

\node[heading,anchor=south]
 at ([yshift=2pt]wr.north)
 {Wiping (GelSight Mini)};

\node[panel] (wf) at ([yshift=-3pt]wr.south)
 {\includegraphics[width=\dimexpr\linewidth-30pt\relax]
 {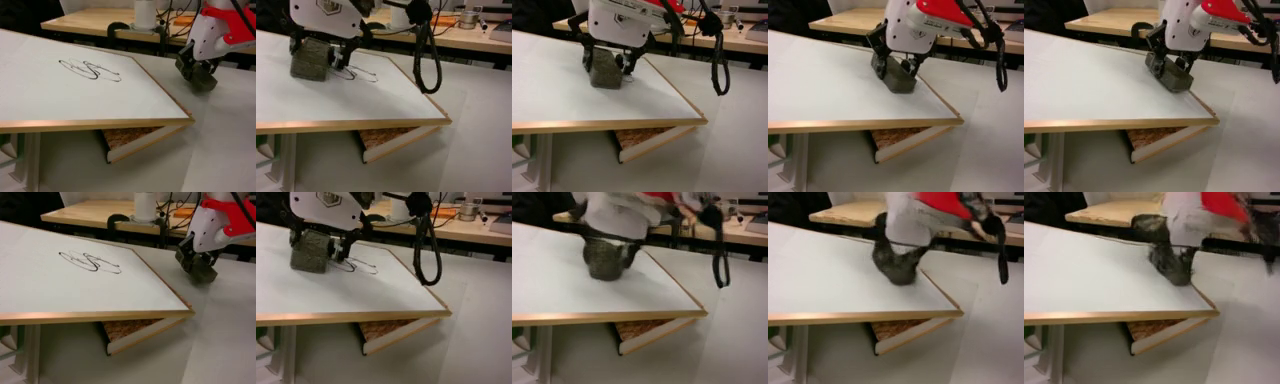}};

\node[panel] (wt) at ([yshift=-3pt]wf.south)
 {\includegraphics[width=\dimexpr\linewidth-30pt\relax]
 {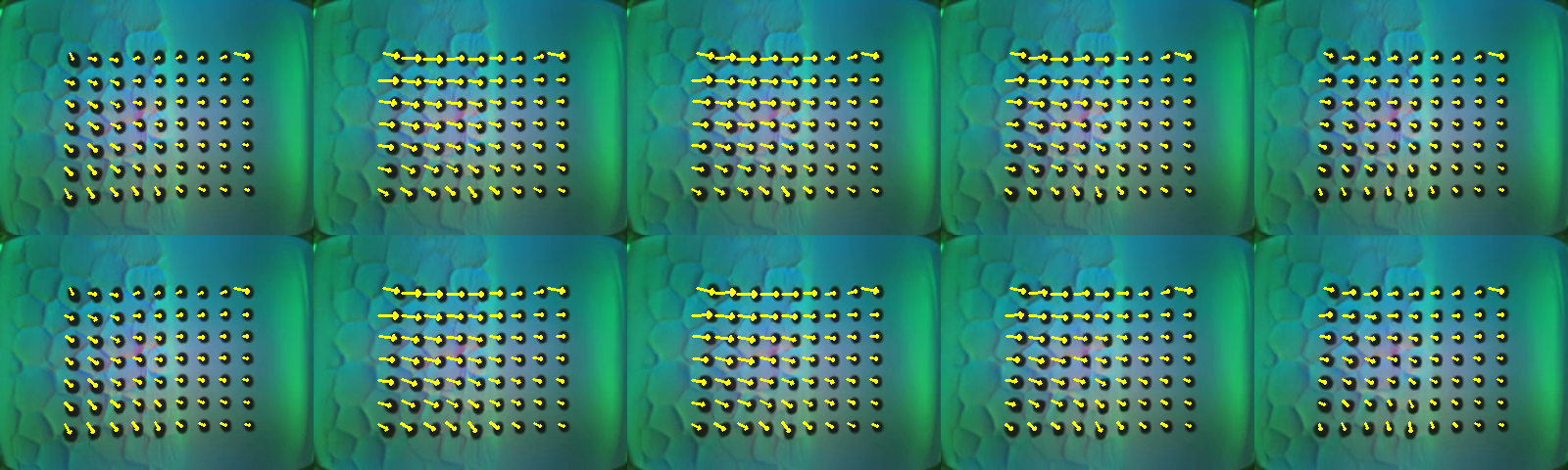}};

\node[heading,anchor=north] (hanoi)
 at ([yshift=-4pt]wt.south)
 {Hanoi Disk Insertion (Xense Photon)};

\node[panel] (hw) at ([yshift=-2pt]hanoi.south)
 {\includegraphics[width=\dimexpr\linewidth-30pt\relax]
 {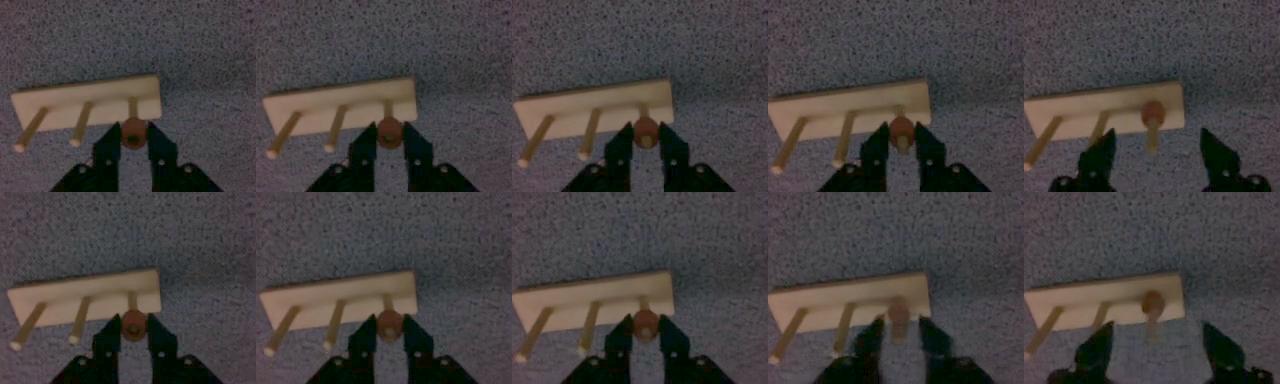}};

\node[panel] (ht) at ([yshift=-3pt]hw.south)
 {\includegraphics[width=\dimexpr\linewidth-30pt\relax]
 {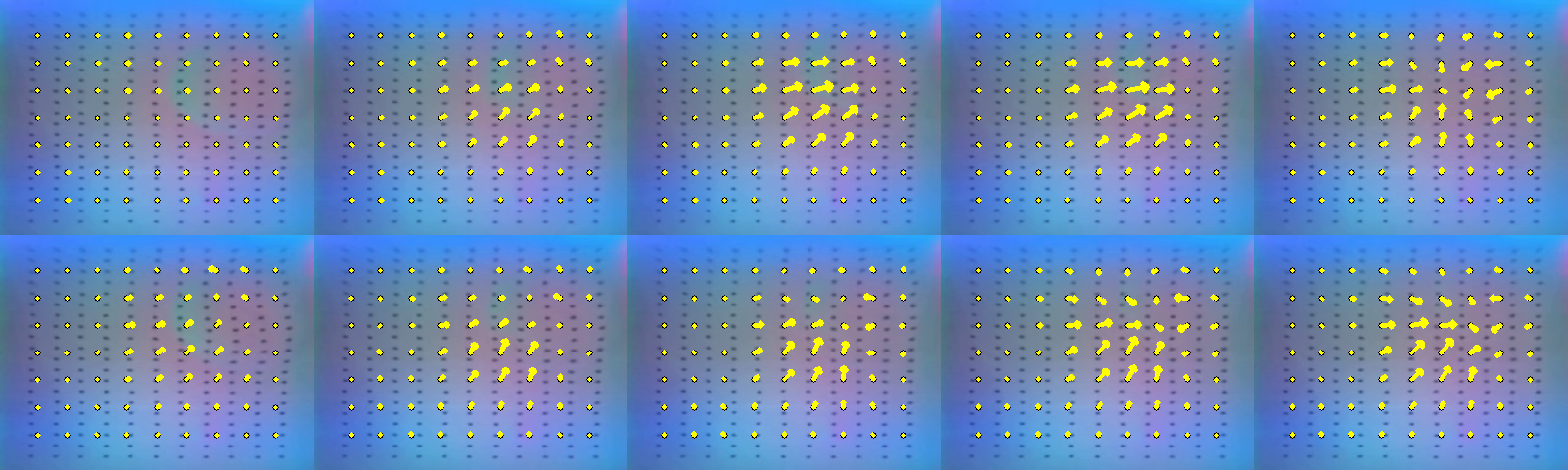}};

\foreach \img/\view/\col in {
  wr/Rear/blue!75,
  wf/Front/orange!85,
  wt/Tactile/purple!80,
  hw/Wrist/teal!75,
  ht/Tactile/purple!80
} {
  \node[lab,anchor=east]
   at ($(\img.north west)!0.25!(\img.south west)$)
   {\rotatebox{90}{GT}};

  \node[lab,anchor=east]
   at ($(\img.north west)!0.75!(\img.south west)$)
   {\rotatebox{90}{\vtam}};

  \node[lab,text=\col]
   at ([xshift=-17pt]\img.west)
   {\rotatebox{90}{\view}};
}

\end{tikzpicture}

\caption{\textbf{Sensory prediction across tactile sensors.}
Each panel compares GT future frames (top) with \vtam predictions
(bottom). Hanoi frames at 0, 0.2, 0.6, 1.0, and 1.4\,s show
peg contact and release within one prediction window; wrist
frames are brightened for display. Arrows illustrate tactile
optical flow (Xense: relative to the last grasp-only frame).}
\label{fig:prediction_wiping}
\vspace{-0.3cm}
\end{figure}

Figure~\ref{fig:prediction_wiping} compares observed (top) and predicted (bottom) future frames for wiping (a--c) and Hanoi Disk Insertion (d--e). \vtam captures scene motion and evolving contact deformation. In the Hanoi sequence, the predicted onset of Xense deformation and subsequent release align with the observed peg-contact and release steps, illustrating how the same predictive framework accommodates tactile sensors with different visual appearances.

\begin{figure}[!t]
\centering
\includegraphics[width=0.99\columnwidth]{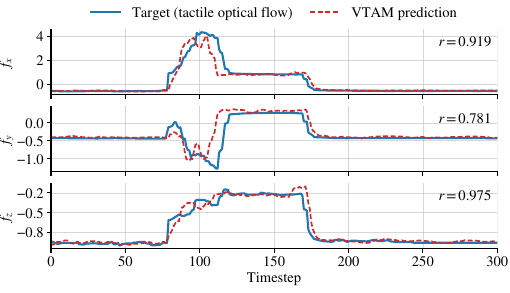}
\caption{\textbf{Flow-force proxy prediction on a held-out chip trajectory.} Predicted components of $F^v$ are compared with targets computed from tactile optical flow ($r$: Pearson correlation).}
\label{fig:force_pred}
\vspace{-0.2cm}
\end{figure}

\noindent\textbf{Flow-force proxy prediction.}
Figure~\ref{fig:force_pred} compares predicted proxy values with targets derived from tactile optical flow on a held-out chip trajectory, showing that predictions track contact onset and release. 

\begin{figure}[t]
\centering
\includegraphics[width=\linewidth]{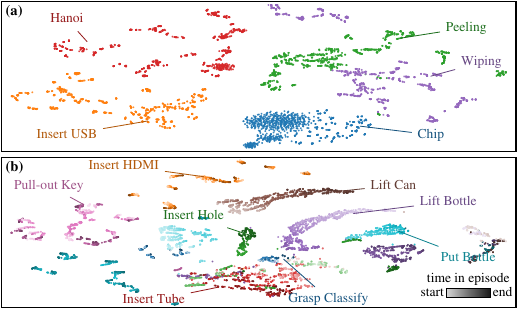}
\caption{\textbf{Zero-shot tactile latents} (t-SNE) of \vtam's frozen video VAE, never trained on touch. (a) Real-robot tactile frames from five tasks (GelSight Mini: chip, peeling, wiping; Xense Photon: Hanoi and Insert USB, used only here), colored by task. (b) UniVTAC tactile frames (8 tasks, 12 episodes each); hue encodes the task and shade (light to dark) normalized time within each episode.\looseness-1}
\label{fig:tactile_tsne}
\vspace{-0.2cm}
\end{figure}

\noindent\textbf{Zero-shot tactile encoding.}
Task-specific linear probes evaluated on held-out episodes distinguish loaded from unloaded frames with 90\%/91\% balanced accuracy and 0.76/0.84 proxy Spearman correlation (real/sim), comparable to raw pixels (88\%/89\%, 0.73/0.83). These results indicate that the frozen encoder preserves contact information across GelSight Mini, Xense Photon, and simulated GelSight without tactile training. 

In Fig.~\ref{fig:tactile_tsne}, the latents separate by task on both real (a) and simulated (b) tactile frames, and within several UniVTAC tasks they are ordered by progress through the episode (b; progress predicted from the embedding on held-out episodes reaches Spearman 0.74 overall and 0.75--0.97 on five of eight tasks). The frozen video VAE thus exhibits structured separation across tasks and, in simulation, within-episode progress, suggesting that its pretrained visual representation already captures interaction-relevant structure. This provides a useful latent space for extending video-pretrained world modeling to tactile observations without large-scale tactile pretraining.\looseness-1

\subsection{Inference Efficiency}
\label{sec:latency}
\begin{table}[!tb]
\centering
\footnotesize
\newcommand{\hd}[2]{\textbf{\begin{tabular}[c]{@{}c@{}}#1\\#2\end{tabular}}}
\caption{\textbf{Inference efficiency.} Latency per action chunk on UniVTAC observations with one NVIDIA L40S (BF16, batch size 1); \vtam is averaged over its eight task models. ``Avg.'': UniVTAC success (Table~\ref{tab:univtac_sim}).}
\label{tab:latency}
\setlength{\tabcolsep}{5pt}
\begin{tabular}{@{}l c c c c c@{}}
    \toprule
    \rowcolor[HTML]{EFEFEF}
    & & & & & \\
    \rowcolor[HTML]{EFEFEF}
    & & & \multicolumn{2}{c}{\textbf{Latency (ms)} $\downarrow$} & \\
    \rowcolor[HTML]{EFEFEF}
    \multirow{-3}{*}{\textbf{Model}} & \multirow{-3}{*}{\hd{Params}{(B)}} & \multirow{-3}{*}{\textbf{\begin{tabular}[c]{@{}c@{}}w/o full\\video\\denoising\end{tabular}}} & \textbf{Eager} & \textbf{Compiled} & \multirow{-3}{*}{\hd{Avg.}{(\%) $\uparrow$}} \\
    \midrule
    LingBot-VA~\cite{li2026lingbotva} & 11.47 & $\times$ & 9{,}380 & 10{,}404 & 31.4 \\
    Fast-WAM~\cite{yuan2026fastwam} & 12.41 & $\checkmark$ & 548 & 127 & 48.0 \\
    Tactile-VLA~\cite{huang2025tactile} & 3.39 & $\checkmark$ & 364 & 230 & 45.4 \\
    FTP-1~\cite{yuan2026ftp} & 3.66 & $\checkmark$ & 345 & 187 & 64.8 \\
    N$_0$-VTLA~\cite{neoteai2026vtla} & 3.81 & $\checkmark$ & 335 & 225 & 83.1 \\
    N$_0$-TWAM~\cite{neoteai2026twam} & 13.59 & $\times$ & 3{,}393 & 5{,}136 & 84.5 \\
    \midrule
    \rowcolor{IllinoisOrange!10}
    \textbf{\vtamc (Ours)} & 7.29 & $\checkmark$ & 588 & \textbf{145} & 83.4 \\
    \bottomrule
\end{tabular}
\end{table}

Table~\ref{tab:latency} reports end-to-end latency per action chunk. \vtam runs one backbone forward pass and 10 Euler steps of the action expert without generating video, taking 588\,ms eager and 145\,ms compiled; on the real robot, five Euler steps take 141\,ms per chunk (about 213 actions/s for 30 executed actions).
Without compilation, \vtam is about 6$\times$ and 16$\times$ faster than N$_0$-TWAM and LingBot-VA, which fully denoise future frames. Compiled, it is comparable to Fast-WAM (127\,ms) while also using touch and reaching higher success (83.4\% vs.\ 48.0\%), and faster than all compiled tactile VLAs (187--230\,ms), matching N$_0$-VTLA (83.1\%) without its tactile pretraining.
\section{Conclusion}
We presented \vtam, an efficient visuo-tactile world-action model for contact-rich manipulation without tactile pretraining. Our central finding is that data-efficient tactile world modeling is feasible: with a suitable architecture and training recipe, a small amount of task-specific fine-tuning data suffices to extend the representations and predictive ability of a pretrained video world model to joint visuo-tactile prediction. Without any large-scale tactile pretraining, \vtam achieves strong contact-rich performance while keeping inference latency low.

\textbf{Future work: scaling toward emergent generalization.} Since a small amount of task-specific fine-tuning already transfers world modeling from video to touch for in-distribution tasks, the main value of large-scale tactile data may lie less in marginal in-distribution gains and more in generalization. A natural next step is to train on highly diverse tactile interaction data spanning tasks, objects, and contact modes, so that the model learns general contact representations, and to study whether new capabilities emerge as data scale and diversity grow, such as zero-shot generalization to unseen contact-rich tasks without further fine-tuning.

\bibliographystyle{plainnat}
\bibliography{main}
\end{document}